# A New Pseudo-color Technique Based on Intensity Information Protection for Passive Sensor Imagery

M. R. Khosravi*, H. Rostami, G. R. Ahmadi, S. Mansouri, and A. Keshavarz

School of Engineering, Persian Gulf University, Bushehr, Iran

*e-mail: scientific@iran.ir

*Abstract*—Remote sensing image processing is so important in geosciences. Images which are obtained by different types of sensors might initially be unrecognizable. To make an acceptable visual perception in the images, some preprocessing steps (for removing noises and etc) are preformed which they affect the analysis of images. There are different types of processing according to the types of remote sensing images. The method that we are going to introduce in this paper is to use virtual colors to colorize the gray-scale images of satellite sensors. This approach helps us to have a better analysis on a sample single-band image which has been taken by Landsat-8 (OLI) sensor (as a multi-band sensor with natural color bands, its images' natural color can be compared to synthetic color by our approach). A good feature of this method is the original image reversibility in order to keep the suitable resolution of output images.

## I. INTRODUCTION

Processing of remote sensing images (satellite images) in order to extract information in different applications is very important. These images are used in various fields such as natural geography, geomorphology and topography engineering (namely, engineers use the images to create the maps). Besides, these satellite images are applicable in climate changes, geophysics, earthquake engineering, soil science, hydrology, jungle engineering, agriculture and desert studies. Remote sensing images have various types including land base, air-borne and satellite images (space-borne). Satellite images can be classified into visible region images, thermal images, radar images and laser images. Each of these images is provided by a specific tool and has its main applications. For example visible region images (optical images) are proper in terms of spatial resolution and understandability, but these images don't give all of the information about the imaged area, so we must use images of other sensors in some usages. Another shortcoming of the visible images that can be mentioned is inability to identify all of the earth surface combinations of the imaged area (due to the passive sensors of visible region). The quality of visible images is low in bad weather conditions and at night. Thermal images are provided in infrared spectrum and normally are used in global warming studies, temperature changes and the relevant studies. The main shortcoming of these images is the low spatial resolution of infrared sensors. Radar images that are commonly provided by active radars have a wide range of applications in geosciences, e.g. object extracting, detecting the buildings and roads, exploration of natural sources and identification of terrain and other corresponding facilities [1].

Table 1. Spectrum of OLI.

| Band No. | Spectral Band Name | Wavelength Range (micro meter) | Spatial Resolution |
|---|---|---|---|
| 1 | Coastal | 0.433 – 0.453 | 30 m |
| 2 | Blue | 0.450 – 0.515 | 30 m |
| 3 | Green | 0.525 – 0.600 | 30 m |
| 4 | Red | 0.630 – 0.680 | 30 m |
| 5 | Nearest IR | 0.845 – 0.885 | 30 m |
| 6 | IR | 1.560 – 1.660 | 30 m |
| 7 | IR | 2.100 – 2.300 | 30 m |
| 8 | Panchromatic | 0.500 – 0.680 | 15 m |
| 9 | Cirrus | 1.360 – 1.390 | 30 m |

### A. Landsat Sensors

For many years a lot of remote sensing satellites have been launched to the space. One or several sensors have been installed on each of these satellites for a special application. Each of these sensors' images is taken in a specific frequency according to its application and provides the information related to that frequency band. Landsat satellites include eight satellites (launched from 1972 to present) and they have been created and launched to the space by NASA and USGS (collaboratively). Today two of them, Landsat-7 and Landsat-8, are active in their orbits. Landsat-7 is the most famous satellite in the series and its multispectral sensor, ETM+, is one of the most famous remote sensing sensors. Landsat-8 has two multispectral sensors. These two sensors, OLI and TIRS, respectively provide remote sensing images in nine and two bands of frequency spectrum. The OLI sensor provides multispectral images and it nearly includes all of the ETM+ bands but with a better quality. The OLI sensor takes images in the visible spectrum and also infrared spectrum, Landsat-8 infrared sensor, TIRS, is a thermal sensor with two infrared bands. The bands of OLI sensor are listed in Table 1. The landsat-8 satellite collects spatial information with a middle resolution quality (range of spatial resolution is about 30 meters). The sensor TIRS has a resolution of 100 meters in both of its bands which are considered as low spatial accuracy bands. The images of Landsat-8 are freely available for all and are in GeoTIFF format, consequently, don't need to geometrical correction due to the previously geo-referencing. According to Table 1, mix of bands 2nd, 3rd and 4th creates a visible image with resolution of 30 meters. The 8th band that has the widest spectrum of visible region (nearly 2/3 of all the spectrum), has spatial accuracy (resolution) of 15 meters (namely, each pixel shows 225 m² on the earth surface). It has a suitable resolution, so this band which has been called "Panchromatic" is known as the band with the highest spatial resolution [2].

## II. RELATED WORKS

In this section, we express some researches done in satellite image processing based on color processing. Samanta *et al.* [4], proposed a satellite image color processing approach by use of HSV [10] color space towards the way that Hue component is used to enhance virtual color of radar images. Daily [5] proposed the algorithm based on brightness intensity quantization in HSV color space. Chuang *et al.* [6] have used the Hue component for image processing without using the virtual colors. Zhiyu *et al.* [7] have proposed a method for processing and creating color by use of curvelet transform in frequency domain where its output has good quality. Jinxiu *et al.* [9] have presented a virtual color coding method for medical images in the RGB color space. Researches of Bovik *et al.* [16]-[17] have been concentrated on the image contents and their effects on visual quality, specially the effect of color in images. They introduced a different metric for evaluating images' quality. In the next sections, we will discuss about their metric in order to quality assessment (QA).

## III. PROPOSED METHOD

To describe an image in the RGB model, each pixel should has three quantities for R, G and B bands, they stand red, green and blue, respectively. The monitor (CRT, LCD, LED and so on) can show the image in color form by having these three quantities. The RGB model is not the only common model; however, it is appropriate for hardware implementation. Fig. 1 shows a three dimensional grid of a color image in general form of the RGB. The image has three gray-scale layers and when they are aggregated together, the result is interpreted as color image. The color image can be easily converted to gray-scale image, but if we wish to convert a black and white image to a color image, we have to use virtual colors or perform color fusion (with using color of another color image). Therefore, an easy way to do the colorization is to create virtual colors with using brightness intensity level quantization algorithms. The aim of colorization of a gray-scale image of 8th band of OLI is to have a better interpretation of spatial resolution without using any lossy spatial domain processing like magnifying, interpolation [11] [15] and non-reversible functions.

Although we can create appropriate color images by use of other spectral bands of OLI sensor, we prefer to create virtual colors based on a pseudo-color process in the image and it is according to this aim that we wish to find a good way for colorizing the intrinsically single-band image, not an original multi-spectral image. In fact, OLI images are used only for testing the proposed scheme. As follows, we describe the proposed method. In order to colorize images, we need to find values of two dimensional matrices R, G and B. Fig. 2 shows a typical curve for brightness intensity level quantization for a gray-scale image. If we classify the brightness levels in three parts and consider each of them to a specified color, the image is seen in color form, but with fake colors. This simple plot shows the main point of the intensity thresholding (or quantization). Practically, there are several algorithms in different applications to do these types of procedures. We do this by use of a color function (Eq. (1)). This function is a color operator which performs lossless processing; it means that the final values of last RGB components of colorized pixel are determined by their brightness intensity and with considering their corresponding pixels in the gray-scale image. So, this function is reversible and guaranties spatial resolution of the colorized image.

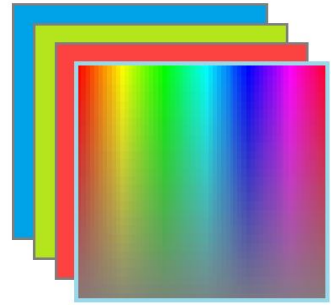

Fig. 1. Color Model.

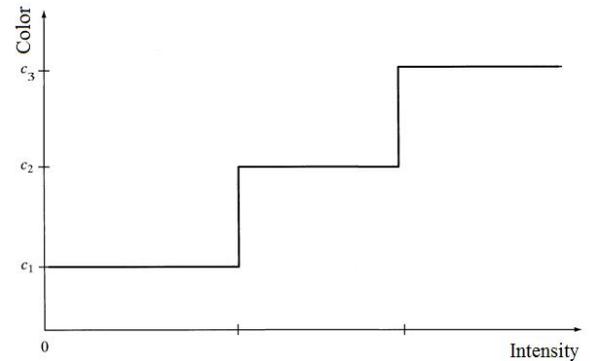

Fig. 2. Brightness level quantization [3].

In Eq. (1), "I" is the value of brightness intensity in gray-scale image and "I' " is the value for the corresponding colorized image, $\alpha$ and $\beta$ are free parameters and we have $0<\alpha<1$, $0<\beta<1$ and $\alpha+\beta<1$. For value selection of $\alpha$ and $\beta$, we can use Table 2 according to the type of the image. The proposed method is reversible and in fact we can create the gray-scale sample by colorized one without any loss of information. We will describe more this feature and its advantages for keeping high resolution data in the simulation result (the fourth section). Eq. (2) shows the reversibility of the proposed method for creating original pseudo-color image band (panchromatic image of OLI in here) based on the colorized image. I'{R}, I'{G}, I'{B} are calculated by I'{R} = 4$\alpha$I, I'{G} = 4$\beta$I and I'{B} = 4(1-$\alpha$-$\beta$)I. After using the mentioned method, we can additionally use some further processes on the colorized image for increasing visual quality of the colorized image. Fig. 3 shows the sequence of this processing.

$$I' = \lim_{|\alpha-\beta| \to 1/4} \begin{cases} 4\alpha I & \forall R \\ 4\beta I & \forall G \\ 4(1-\alpha-\beta)I & \forall B \end{cases} \quad (1)$$

$$I = \frac{I'\{R\} + I'\{G\} + I'\{B\}}{4} \quad (2)$$

Table 2. Color selection.

| Region | Color Region | Values of α and β |
|---|---|---|
| 1 | Red - Yellow | α>β>(1-α)/2 |
| 2 | Yellow - Green | β>α>(1-β)/2 |
| 3 | Green - Cyan | β>(1- α)/2 , α<1/3 |
| 4 | Cyan - Blue | (1-α)/2>β>α |
| 5 | Blue - Magenta | (1-β)/2>α>β |
| 6 | Magenta - Red | α>(1-β)/2 , β<1/3 |

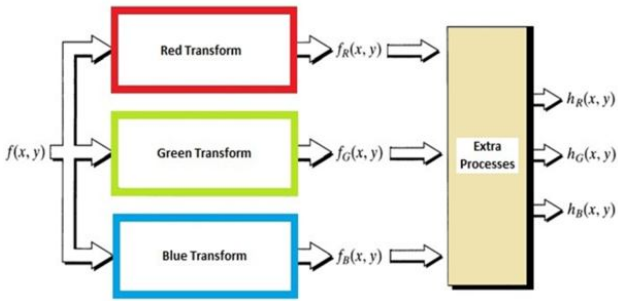

Fig. 3. Pseudo-color processing procedure.

## IV. SIMULATION RESULTS

Two software tools are used in this paper to do simulation work, MATLAB for main processing and color creation, ENVI 4.7 for pre-processing and final processes. Preprocessing is carried out to remove noises and radiometric corrections (if they are essential). In here, images do not need to the geometrical corrections due to geo-referencing. Images used in this paper are related to Doroodzan dike's lake in Fars province of Iran. Fig. 4.b shows a visible image with 30 meters resolution that created by combining images of bands 2nd, 3rd and 4th of the OLI sensor. Fig. 4.a shows single-band image of its 8th band which gas 15 meters resolution. Fig. 4.c and Fig. 4.d show the colorized images from Fig. 4.a by the represented algorithms in [4] [20] and [5].

Algorithm in [4] which has been shown in Eq. (6) and Table 6 is a color enhancer technique that operates on the colorized image created by the method in [20] (the process in [20] is like Fig. 2). Although image of Fig. 4.d has a good colorfulness, it loses the details, especially depth information. In appendix, implementation tools for the methods in [5] and [20] will be described. To colorize an image, it is just needed to read values of α and β from Table 2. Two images of different color regions are show in Fig. 5.a and Fig. 5.b, and the values of α and β are represented in Table 3. The shown images are in blue-magenta and cyan-blue regions, respectively. In first step for performance evaluation, performance of the proposed method is visually evaluated and also by an error criteria.

And then, the second step is about the extra processes. This step is very important and usually implemented by software blocks. Now, we are going to evaluate the performance of the proposed method based on the error criterion. To do this, we compare the colorized images to the image with natural color (Fig. 4.b). Colorized images of Fig. 4.c (by Samanta/Otsu methods) and Fig. 4.d (by Daily method) are compared to the image of Fig. 5.a that colorized by the proposed method. The amount of error for each colorized process is calculated by Root-Mean-Square-Error (RMSE) that is written as Eq. (3).

$$RMSE = \left( \frac{1}{MN} \sum_M \sum_N (f - f')^2 \right)^{1/2} \quad (3)$$

This exam is based on Human Visual System (HVS) and is a special exam which is only used in the specific example related to evaluation of the proposed method because of in other images, reference image does not exist. However, the proposed method is not only for a specific satellite sensor. Error values are calculated and listed in Table 4. Normalized RMSE (S), S stands the saturation error, represents the amount of colorized error for color components in the HIS color model of the gray-scale image. In fact, RMSE (S) shows image depth (altitude) information error in the colorized satellite images.

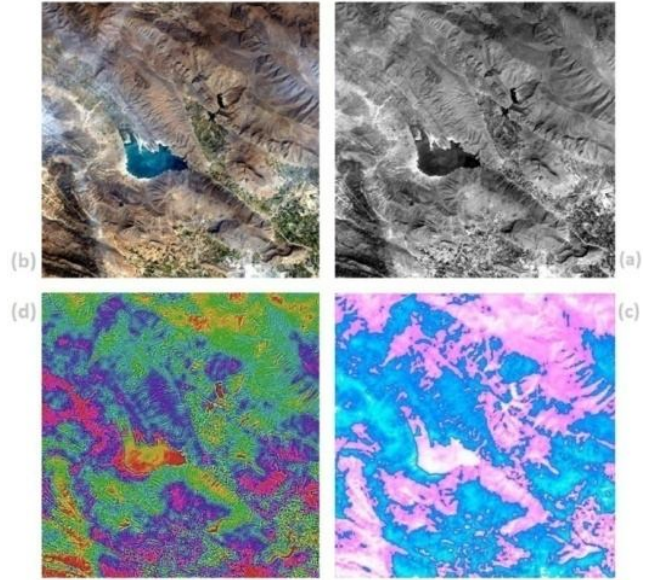

Fig. 4. Part (a) is gray-scale, (b) shows original color image (natural), (c) is output of Otsu/Samanta method (virtual) and (d) is output of Daily method (virtual).

Table 3. Values of α and β in Fig. 5.

| Output | Technique (Proposed Approaches) | α and β |
|---|---|---|
| Fig. 5.a | Proposed Color Function (CF) | β= 0.15 / α=0.38 |
| Fig. 5.b | Proposed Color Function (CF) | β= 0.33 / α= 0.16 |
| Fig. 5.c | Proposed CF + HM | β= 0.15 / α=0.38 |
| Fig. 5.d | Proposed CF + GLPF | β= 0.15 / α=0.38 |

Table 4. Results for saturation error.

| Output | Technique | Normalized RMSE |
|---|---|---|
| Fig. 5.a | Proposed Method | 0.3692 |
| Fig. 4.c | Otsu/Samanta Method | 0.5744 |
| Fig. 4.d | Daily Method | 0.4465 |

Evaluation of outputs shows that although hue error of the proposed method is more than the others (due to use of two color intervals only); it has a better visual quality for the colorized image's details. Saturation error confirms this fact. It can be seen that the colorized images by [4] and [5] have more error (due to use a fixed color to an interval of brightness intensity). In the following parts, we evaluate the effect of a post-processing on the visual quality. Fig. 6 (3 curves in a diagram) shows the histogram of the color image of Fig. 4.b and its sub-bands (R, G and B). So one of the post-processing that can be used to improve final results is to correlate the colorized image histogram with visible image with using histogram matching, Fig. 5.c shows its output. This process uses some external information; therefore it is a kind of image fusion [12-14]. In many applications, we cannot use fused information because the information is not available, however, we do it because it causes a logical conclusion. If fusion is possible, it is the best process for colorizing, for example Choi et al. [8] has proposed a method for creation of good quality, but in single-band images this is not possible.

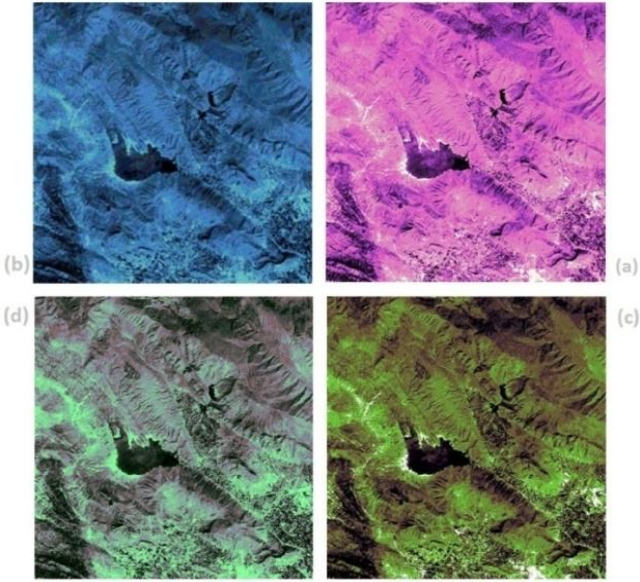

Fig. 5. Colorized images; Parts (a) and (b) are outputs of the proposed method without any post-processing (only our color function), (c) is the output of (a) which has been processed under Histogram Matching (HM) and (d) shows it under Gaussian LPF (GLPF).

We emphasis on this point again, that our purpose is not only Landsat images and it consists of other satellite images with low or middle resolution and the single-band sensor. Fig. 5.d is the output of the colorized image under our method (Fig. 5.a) and after passing through a typical Gaussian low pass filter (GLPF was implemented by ENVI 4.7); this image is highly similar to Fig. 5.c, however, it has no need to any type of fused information. GLPF is a frequency domain filter that its equal mask in spatial domain is shown in Eq. (4). It is clear that this mask represents a low pass filter.

$$\begin{bmatrix} 0.0007 & 0.0256 & 0.0007 \\ 0.0256 & 0.894834 & 0.0256 \\ 0.0007 & 0.0256 & 0.0007 \end{bmatrix} \quad (4)$$

This last process shows due to the fact that histogram of the color bands in the color image commonly is similar to Gaussian distribution, so outputs of histogram matching with a real optical image and a Gaussian filtering are the same. Consequently in the cases that fusion is not possible; the approximate process for histogram matching is Gaussian filtering. This designed experiment in this paper only uses Landsat images for representation of this fact. Results of SSIM metric are co-directive with this fact too, and Table 5 shows the experimental results. SSIM is new metric which was proposed by Wang *et al.* in 2004 [16] and is an excellent selection in terms of HVS. It has been represented in Eq. (5) [18]. In fact, SSIM is a popular metric for image quality assessment, which compares local patterns of pixels from three components luminance, contrast, and structure. For two images $x$ and $y$, the SSIM value is evaluated as Eq. (5), where $m_x$ and $m_y$ represent the mean values, $\sigma_x$ and $\sigma_y$ denote the standard deviations, and $\sigma_{xy}$ is the cross correlation of the mean-shifted images $x - m_x$ and $y - m_y$. The constants $C_1$, $C_2$, and $C_3$ are some constants and are not usually zero. In practice, the comparisons are computed within a weighting window which moves pixel by pixel over the entire image to obtain a spatially varying SSIM map.

$$SSIM = \frac{(2u_x u_y + C_1)(2\sigma_x \sigma_y + C_2)(\sigma_{xy} + C_3)}{(u_x^2 + u_y^2 + C_1)(\sigma_x^2 + \sigma_y^2 + C_2)(\sigma_x \sigma_y + C_3)} \quad (5)$$

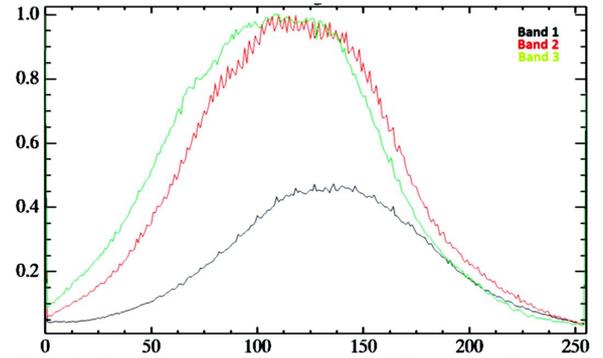

Fig. 6. Histogram of the natural RGB image (Fig. 4.b).

This quality map can reject the localized quality assessment, and the global average value of SSIM map is SSIM value. For image fusion evaluation [19], SSIM is used to measure the similarity relationship between source images or between fused image and one source. The main drawback is its lack of information for HVS. So for computation of SSIM, we must compute five input values consist of two means, two standard deviations and images covariance. SSIM shows these two images have about 78% structural similarity. An important point in the image colorization is the reversibility of colorization algorithm. It means that we can create the main gray-scale version of the colorized image from it. Importance of this point is the fact that we can achieve all information with high resolution whether been preserved or not. So non-reversible algorithms that some of them have high color quantization error reduce image resolution and lose depth information. It means that they remove some information. The details can be shown in brightness intensity of the colorized image. Therefore, this is one of the vital characteristic of the proposed method in comparison to other usual techniques.

## V. CONCLUSION

Due to the limited access to freely available satellite images with high spatial resolution, it is very important to use the lower spatial resolution images that can easily be accessed. The propose method in this article for colorizing the highest resolution band in OLI sample can be used in other sensors of RS satellites. It is possible to obtain a low saturation error when using a reversible function for color generation, however, post-process may be non-reversible.

Table 5. Structural similarity.

| Color Band | SSIM |
|---|---|
| Red | 0.5563 |
| Green | 0.7302 |
| Blue | 0.4163 |
| Average | 0.5676 |

## VI. APPENDIX

Daily method [5] was implemented by ENVI 4.7. For this work, we use a special tool in the software that ENVI help has more information about it. Otsu method [20] that is a simple technique for pseudo-color processing based on segmentation and intensity quantization, was implemented by MATLAB R2010a (image processing toolbox). Samanta method [4] is a post-process (final process) for output of Otsu method where Eq. (6) and Table 6 show details of this method.

$$Hue = 60° \left( m + n \frac{M-L}{H-L} \right) \quad (6)$$

where $L \leq M \leq H$ and $m \in \{0, 2, 4, 6\}$, $n = \pm 1$

Table 6. Parameters for Eq (6).

| Color Regions | Parameters |
|---|---|
| Red - Yellow | m = 0 , n = 1, B<G<R |
| Yellow - Green | m = 2 , n = -1, B<R<G |
| Green - Cyan | m = 2 , n = 1, R<B<G |
| Cyan - Blue | m = 4 , n = -1, R<G<B |
| Blue - Magenta | m = 4 , n = 1, G<R<B |
| Magenta - Red | m = 6 , n = -1, G<B<R |

Representation for Gaussian filter is as below. $D_0$ is a free parameter in Eq. (7) and $(u,v)$ is two independent variables in the frequency domain. In this paper, we use low pass filter (LPF) condition.

$$H(u,v) = \begin{cases} e^{-D^2(u,v)/2D_0} & LPF \\ 1 - e^{-D^2(u,v)/2D_0} & HPF \end{cases} \quad (7)$$

## AUTHOR'S PROFILE

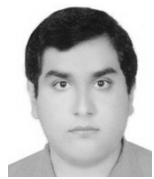

**Mohammad Reza Khosravi**
received his B.Sc. degree from Shiraz University, Iran and the M.Sc. degree from Persian Gulf University, Iran and both are in Electrical Engineering. His interests contain wireless ad-hoc and sensor networks, satellite remote sensing and image processing.

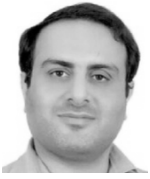
**Habib Rostami**
received B.Sc., M.Sc., and Ph.D. in Computer Engineering from Sharif University of Technology, Iran. Now he is an Assistant Professor at Persian Gulf University, Bushehr, Iran. His interests include computer networking and traffic engineering, neural networks, data mining, statistical pattern recognition, optimization and soft computing, and graph theory.

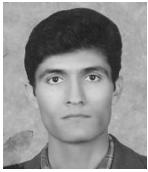
**Gholam Reza Ahmadi**
received his B.Sc degree in Computer Engineering from the University of Tehran, Fanni Faculty, in 2000, and has received his M.Sc. degree in Information Technology Engineering from Amir Kabir University, 2010. He is working in Persian Gulf University (Jam Branch) now. He has taught in the areas of computer and network and his research interests include data mining, and image processing.

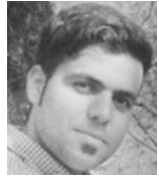
**Suleiman Mansouri**
received the B.Sc. degree in Electrical Engineering (Communications) from Persian Gulf University (PGU) in 2014 and he is now working on his M.Sc. thesis at Persian Gulf University. His research interests include image processing, compression and coding and remote sensing systems.

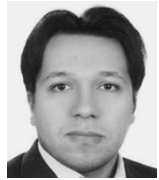
**Ahmad Keshavarz**
received B.Sc. degree from Shiraz University in 2001, Iran and M.Sc. degree from Tarbiat Modares University (TMU) in 2004, and Ph.D. degree from TMU in 2008, and all in Communication Engineering. His research interests include remote sensing image processing and statistical pattern recognition.